\documentclass[conference]{IEEEtran}
\IEEEoverridecommandlockouts
\usepackage{cite}
\usepackage[nodisplayskipstretch]{setspace}
\usepackage{amsmath,amssymb,amsfonts}
\usepackage{mathtools}
\usepackage{url}
\usepackage{hyperref}
\usepackage{xcolor}
\usepackage{breqn}
\usepackage{graphicx}
\usepackage{textcomp}
\usepackage{grffile}
\usepackage{xcolor}
\usepackage{color}

\usepackage{amsmath}
\usepackage{amsfonts}
\usepackage{amssymb}

\usepackage{algorithm,algpseudocode}

\newcommand{\beq}{\begin{equation}}
\newcommand{\eeq}{\end{equation}}
\newcommand{\beqn}{\begin{eqnarray}}
\newcommand{\eeqn}{\end{eqnarray}}
\newcommand{\beqno}{\begin{eqnarray*}}
\newcommand{\eeqno}{\end{eqnarray*}}
\newcommand{\bma}{\begin{displaymath}}
\newcommand{\ema}{\end{displaymath}}
\newcommand{\bnu}{\begin{enumerate}}
\newcommand{\enu}{\end{enumerate}}
\newcommand{\bce}{\begin{center}}
\newcommand{\ece}{\end{center}}
\newcommand{\btb}{\begin{tabular}}
\newcommand{\etb}{\end{tabular}}
\usepackage{graphbox}

\def\BibTeX{{\rm B\kern-.05em{\sc i\kern-.025em b}\kern-.08em
    T\kern-.1667em\lower.7ex\hbox{E}\kern-.125emX}}
    
\begin{document}

%\title{Semi-supervised Learning of Human Actions for Smart Health Monitoring\\
%\title{Sensor-based Human Activity Recognition in Smart Health Monitoring System: A Semi-supervised Machine Learning\\
%\title{Big Data Management for Secured Smart Healthcare System: A Machine Learning Framework\\
\title{Machine Learning-based Framework for Smart Healthcare System
%{\footnotesize \textsuperscript{*}Note: Sub-titles are not captured in Xplore and
%should not be used}
%\thanks{Identify applicable funding agency here. If none, delete this.}
}

\author{Abrar Zahin, Le~Thanh~Tan,~\IEEEmembership{Member,~IEEE} and Rose~Qingyang~Hu,~\IEEEmembership{Fellow,~IEEE} 
\thanks{A.~ Zahin, L.~T.~Tan and R.~Q.~Hu are with the Department of Electrical and Computer Engineering, Utah State University, Logan, Utah 84322-4120, USA. 
Emails: abrarzahin303@gmail.com; \{tan.le, rose.hu\}@usu.edu.}\\

%\and
%\IEEEauthorblockN{2\textsuperscript{nd} Given Name Surname}
%\IEEEauthorblockA{\textit{dept. name of organization (of Aff.)} \\
%\textit{name of organization (of Aff.)}\\
%City, Country \\
%email address}
%\and
%\IEEEauthorblockN{3\textsuperscript{rd} Given Name Surname}
%\IEEEauthorblockA{\textit{dept. name of organization (of Aff.)} \\
%\textit{name of organization (of Aff.)}\\
%City, Country \\
%email address}
%\and
%\IEEEauthorblockN{4\textsuperscript{th} Given Name Surname}
%\IEEEauthorblockA{\textit{dept. name of organization (of Aff.)} \\
%\textit{name of organization (of Aff.)}\\
%City, Country \\
%email address}
%\and
%\IEEEauthorblockN{5\textsuperscript{th} Given Name Surname}
%\IEEEauthorblockA{\textit{dept. name of organization (of Aff.)} \\
%\textit{name of organization (of Aff.)}\\
%City, Country \\
%email address}
%\and
%\IEEEauthorblockN{6\textsuperscript{th} Given Name Surname}
%\IEEEauthorblockA{\textit{dept. name of organization (of Aff.)} \\
%\textit{name of organization (of Aff.)}\\
%City, Country \\
%email address}
}

\maketitle

\begin{abstract}
In this paper, we propose the novel framework for the smart healthcare system, where we employ the compressed sensing (CS) and the combination of the state-of-the-art machine learning based denoiser as well as the alternating direction of method of multipliers (ADMM) structure. This integration dramatically simplifies the software implementation for the low-complexity encoder, thanks to the modular structure of ADMM. Furthermore, we focus on detecting fall down actions from image streams. 
Thus, our primary purpose is to reconstruct the image as visibly clear as possible and hence it helps the detection step at the trained classifier. For our efficient smart health monitoring framework, we employ the trained binary convolutional neural network (CNN) classifier for the fall-action classifier, because this scheme is a part of surveillance scenario. In this
scenario, we deal with the fall-images, thus, we compress, transmit and reconstruct the fall-images.
Experimental results demonstrate the impacts of network parameters and the significant performance gain of the proposal compared to traditional methods.
\end{abstract}


\section{Introduction}
Typically, a surveillance network is consisted of wireless camera nodes, which generates an extensive amount of images.
These images are then transmitted to the processing center, which is responsible for detecting any anomalies \cite{rob_cS_image, rob_cS_image_1}. 
Due to limited resource and computational capability, it becomes unaffordable for a camera node to maintain the stream of images for a very long time. 
For example, according to IEEE 802.15.4, highest data rate can be achieved at physical layer is 250kbps at the 2.4GHz band, which is too low to stream images in a good enough quality for real time scenarios like surveillance. 
Thus, reducing the magnitude of transmitted samples can be beneficial for saving energy and combating overwhelming data bandwidth. 
Furthermore, it is crucial to design an efficient image compression and transmission scheme to develop a prolonged camera sensor networks.

There exists several challenges, while designing such schemes with traditional JPEG and JPEG 2000. 
The traditional approaches can achieve excellent compression performance, however, their computational complexities makes them unsuitable for resource-constrained camera sensor nodes \cite{rob_cS_image_1}. 
Note that, the encoder in a traditional surveillance network normally runs on the low-powered nodes, while the decoder runs on the powerful computer. 
Thus, it is desirable to shift the computation burden to the decoder to implement the low-complexity encoder. 
Furthermore, our desired scheme must be robust to packet losses in a wireless channel. 
Traditionally, images are divided into multiple packets before transmitting via a wireless channel. 
Thus, the missing of excessive or important packets not only degrades the reconstructed image quality but also jeopardizes the surveillance action.
This consequently makes other transmitted packets meaningless. 
Recently, compressive sensing (CS) based image coding has been investigated as an effective solution to address the packet losses for image transmission \cite{rob_cS_image, rob_cS_image_1}. 
The key idea of this solution is that the quality of reconstructed image only depends on the number of CS measurements, not on which of the measurements that are received.

In the literature, the most popular image compression schemes, e.g. Discrete Cosine Transform and Discrete Wavelet Transform  are based on the transform coding \cite{dct_dwt_not_good}. 
To reduce bandwidth and energy usage, the high-energy transform coefficients, which represent the most important features only, would be transmitted \cite{discard_imp_ones}. 
Another improvement is using the distribution image compression schemes \cite{dis_comp_sche_1, dis_comp_sche_2}, where they could reduce the computation cost and the energy consumption of each sensor node by exploiting the parallel processing and by dividing the workload among individual nodes. 
However, the active workload sharing and the dynamic cooperation in these methods can significantly increase the communication overhead \cite{commu_overhead}. 
Hence, these algorithms must use the necessary error correction mechanisms (e.g., Forward Error Correction, multi-path transport and automatic repeat request) to address the possible packet loss during transmission.
These additive mechanisms can further complicate the implementation at the sensor nodes.
To address these arising issues, the alternative direction method of multipliers (ADMM) in \cite{ADMM_Intro, ADMM_Intro_1} is well suited to distributed convex optimization and is particularly fitted to the large-scale reconstruction problem.

The contributions of our paper can be summarized as follows.
We propose the novel framework (CS-ADMM) for the smart healthcare system by exploiting the CS and the alternating direction of method of multipliers (ADMM) structure. 
This integration dramatically simplifies the software implementation for the low-complexity encoder, thanks to the modular structure of ADMM. 
This robust structure can play a vital role in reducing the energy consumption in various IoT environment without compromising the signal quality. 
Furthermore, we focus on detecting fall down actions from image streams. 
Thus, our primary purpose is to reconstruct the image as visibly clear as possible and hence it helps the detection step at the trained classifier. 
If the degradation is not graceful detecting actions properly at the decoder end, then it might raise a significant mis-detections. 
For our efficient smart health monitoring framework, we employ the trained binary CNN classifier for the fall-action classifier, because this scheme is a part of surveillance scenario. 
In this scenario, we deal with the fall-images, thus, we compress, transmit and reconstruct the fall-images.

\section{System Model}

\begin{figure*}[ht]
\centering
\includegraphics [width=120mm]{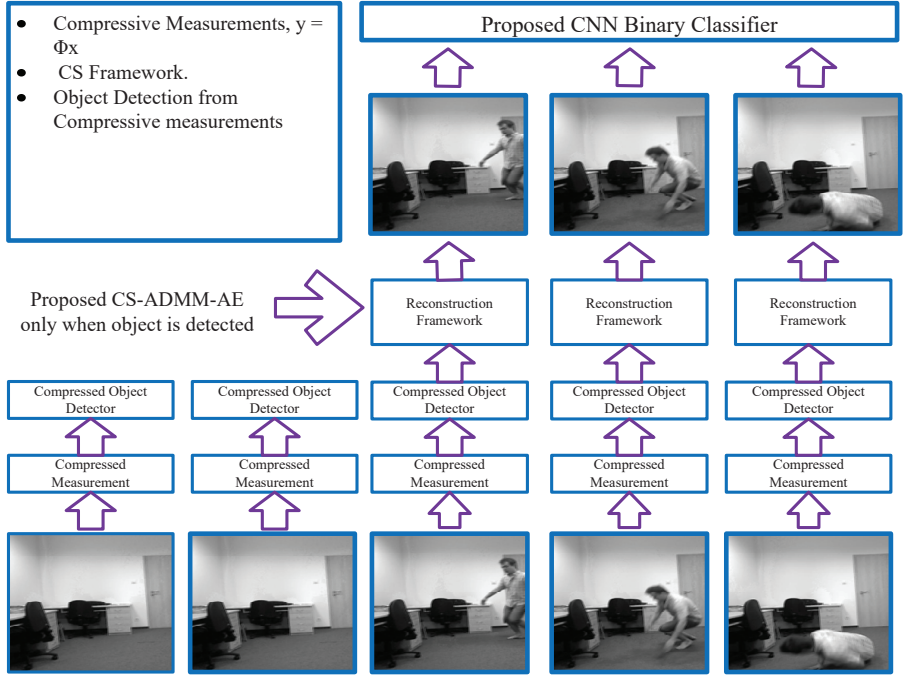}
\caption{An example of sequence of video frames. Here, object is appeared in last three frames. Thus, after detecting \textit{frames with object} from the compressive measurements, our proposed framework reconstructs only those frames and send them to the classifier.}
\label{fall_reco_class}
\end{figure*}

\subsection{CS Recovery of Image} 

The introduction of CS \cite{csintro1, csintro2, Thanh10, Tan16, csintro3} is a major breakthrough in signal processing community. 
CS is basically used for the acquisition of signals, which are either sparse or compressible. 
Sparsity is the inherent property of those signals for which, whole of the information contained in the signal can be represented only with the help of few significant components, as compared to the total length of the signal. 
Similarly, if the sorted components of a signal decay rapidly obeying power law, then these signals are called compressible signals.
A signal can have sparse representation either in original domain or in some transform domains like Fourier transform, cosine transform, wavelet transform, etc. 
A few examples of signals having sparse representation in certain domain are 1) natural images, which have sparse representation in wavelet domain, 2) speech signal, which can be represented by fewer components using Fourier transform, and 3) medical images, which can be represented by using Radon transform. 
%In this paper, we propose a novel approach, where compressed images has been denoised by cDAE exploiting Plug-and-Play framework (P\&P) \cite{pnp}, which is a variant of ADMM. 

\subsection{Fall Action Detection and Alarm}

Finally, our goal is to detect whether the object in the frame (i.e. the person) has fallen or not at the decoder side (i.e. cloud or other monitoring centers). 
It can be easily done by using an already trained classifier, which performs on reconstructed frames separately.

\section{Compressive Sensing-based Image Reconstruction and Fall Detection}

In the following sections, we present our proposed CS-based frame reconstruction applied to the images. 

\subsection{CS-based Fall-Frame Reconstruction}
\label{csimage}

In this subsection, we mathematically formulate the procedures of CS acquisition and reconstruction. 
Let us denote our original fall-frame by a matrix, $X \in \mathbb{R}^{\sqrt{N} \times \sqrt{N}}$ .
Its corresponding vectorized form and its measurements are $x \in \mathbb{R}^{N}$ and $y \in \mathbb{R}^{M}$, respectively.
Note that $M << N$ and the ratio $M/N$ is called sampling rate or sub-rate. 
Then, the signal acquisition problem is formulated as the linear projection 
\beqn
\label{basiccseq}
&& y = \Phi x,
\eeqn
where $\Phi \in \mathbb{R}^{M \times N}$ is a projection matrix. 
For a given sampling rate, $\phi$ is usually constructed by generating a random Gaussian matrix and then orthogonalizing its rows, i.e. $\Phi \times \Phi ^{T} = I$. 
According to the CS theory, if our input signal $x$ meets the sparsity requirement, it would be robustly reconstructed from the measurements. 
Directly computing $x$ from (\ref{basiccseq}) is an under-determined problem due to the lower dimension of the measurement $y$. Thus, this inverse problem can only be solved satisfactorily by adopting some sort of regularization (or prior information in Bayesian inference terms).
\beqn
\label{cs2}
&& \min_{x} \, g(x) \,\,\, s.t \,\,\, y = \Phi x.
\eeqn

In (\ref{cs2}), $g(x)$ represents a prior model, which depicts some intrinsic characteristics of the original signal. 
Basically, the prior model gives the preferences to a solution with desirable properties, which determines the reconstruction efficiency.
Rather than (\ref{cs2}), most state-of-the-art CS reconstruction algorithms consider the following unconstrained problem
\beqn
\label{cs3}
&& \hat{x} = \min_{x} \, ||\phi x - y||^{2}_{2} \,\, + \,\lambda g(x),
\eeqn
where $\lambda$ ($\lambda \in \mathbb{R}_{+}$) is the regularization parameter.

\subsection{Optimizing with ADMM}

\begin{figure*}[ht]
\centering
\includegraphics [width=120mm]{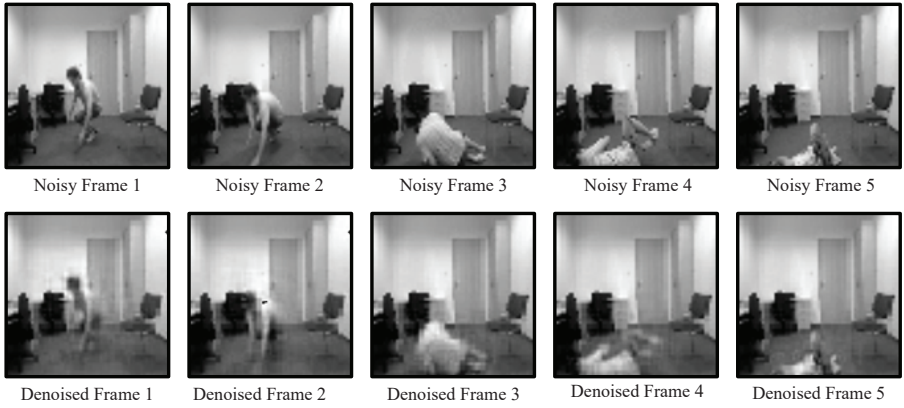}
\caption{Denoised performance results}
\label{denoised_re}
\end{figure*}

\begin{figure*}[tp]
\centering
\includegraphics [width=120mm]{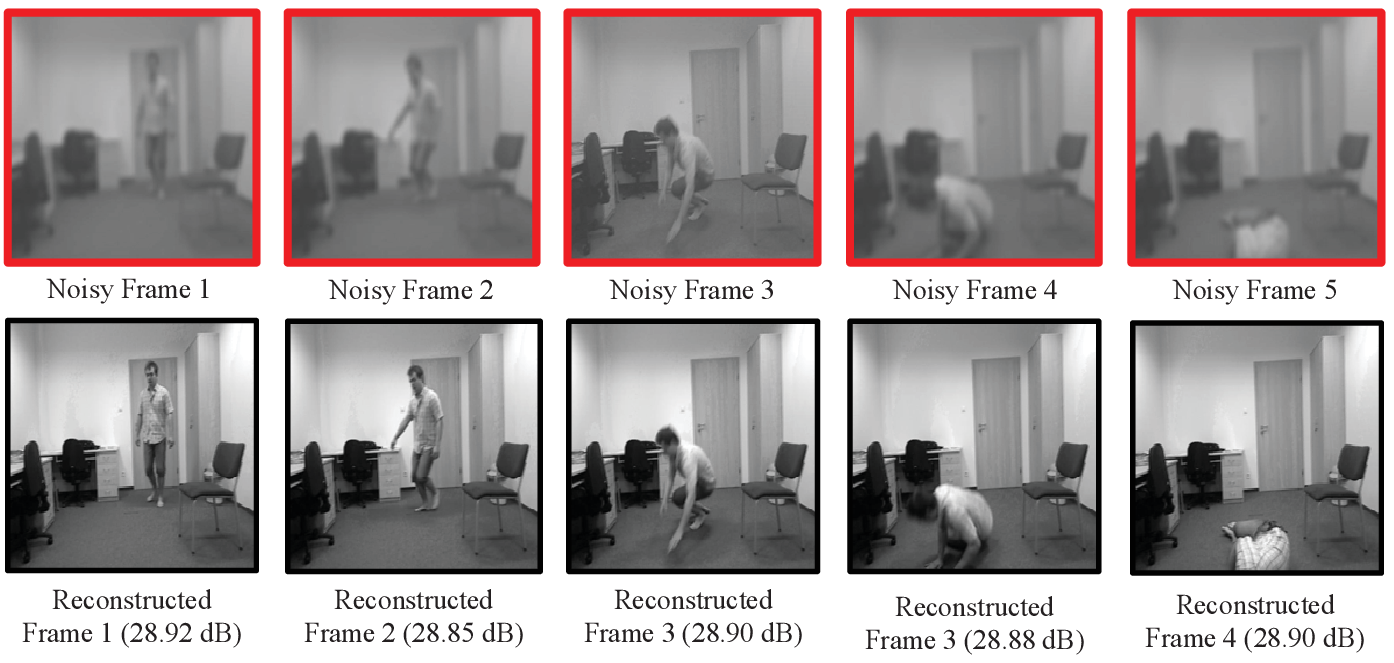}
\caption{Frame reconstruction performance of the proposed CS-ADMM framework. First row presents the noisy frames, while the second row illustrates the restored ones.}
\label{reconstructed_re}
\end{figure*}

In this paper, we tackle the unconstrained optimization problem for reconstructing fall frames (refer to (\ref{cs3})) by leveraging the concepts of Alternating Direction Method of Multipliers (ADMM) .

\subsubsection{Alternating Direction Method of Multipliers (ADMM)}

In order to facilitate the discussions in the following optimization section, this subsection briefly introduces the ADMM, which has become a workhorse for a variety of problems in the form of (\ref{cs3}). 
The basic idea of ADMM is to replace a constrained optimization problem by a series of unconstrained problems and to add a penalty term to the objective. 
These are done by exploiting the variable splitting technique, which is then followed by invoking the augmented Lagrangian method.

Let rewrite the optimization problem (\ref{cs3}) in the following format by considering $f(x) =||\phi x - y||^{2}_{2}$
\beqn
\label{1}
&& \hat{x} = \min_{x} \, f(x) \, + \,\lambda g(x).
\eeqn
The idea of ADMM is to convert (\ref{1}), an unconstrained optimization problem, into a constrained one, given as
\beqn
\label{2}
&& \hat{x},\hat{v} = \min_{x,v} \, f(x) \, + \,\lambda g(v), \,\,\,\, s.t \,\,\, x=v.
\eeqn 
The logic behind the variable splitting is explained as follows.
By decoupling $f$ and $g$, solving the constrained problem (\ref{2}) is easier than solving its unconstrained counter-part (\ref{1}). 
So the optimal solution of (\ref{2}) is denoted by
\beqn
p^{*} = inf \{f(x) + g(v))\,\,\, | \, x = v \}.
\eeqn

Using the method of multipliers, we form the following augmented Lagrangian with penalty parameter $\rho$,
\beqn
\label{3}
\mathcal{L}_{\rho} (x,v, \vartheta) = f(x)+ \lambda g(v)+ \vartheta^{T}(x-v)+ \frac{\rho}{2} ||x-v||_{2}^{2}.
\eeqn
Here, $f(x)+ \lambda g(v)+ \vartheta^{T}(x-v)$ is the \textit{Lagrangian} part, whereas $\frac{\rho}{2} ||x-v||_{2}^{2}$ is the \textit{Augmented} part.
It can be seen that the quadratic penalty in the augmented Lagrangian destroys the separability of the Lagrangian. 
Therefore, it is impossible to minimize the augmented Lagrangian by separately minimizing over the variables $x$ and $v$. 
To solve this problem, $x$ and $v$ are updated in an alternating or sequential fashion, which accounts for the term \textit{alternating direction}. 
Thus, ADMM consists of following iterations
\beqn
\label{4} && x^{(k+1)} := \underset{{x \in \mathbb{R}^{n}}}{\mathrm{argmin}}\, f(x)\, + \frac{\rho}{2} ||x- \tilde{x}^{(k)}||_{2}^{2},  \\
\label{5} && v^{(k+1)} := \underset{{v \in \mathbb{R}^{n}}}{\mathrm{argmin}}\, \lambda g(v) + \frac{\rho}{2} ||v- \tilde{v}^{(k)}||_{2}^{2}, \\
\label{6} && \bar{\vartheta}^{(k+1)} \,\, \leftarrow \,\, \bar{\vartheta}^{(k)} + (x^{(k+1)}-v^{(k+1)}),
\eeqn
where $\rho$ is the augmented Lagrangian parameter and $\rho > 0$. 
$\bar{\vartheta}^{(k)}$ is the scaled Lagrange multiplier and $\bar{\vartheta}^{(k)} \overset{\Delta}{=} (1/\rho)\vartheta^{(k)}$.
Also, we have $\tilde{x}^{(k)} \overset{\Delta}{=} v^{(k)} - \bar{\vartheta}^{(k)}$ and $\tilde{v}^{(k)} \overset{\Delta}{=} x^{(k+1)} - \bar{\vartheta}^{(k)}$. 
Under mild conditions, e.g. when both $f$ and $g$ are closed, proper and convex and if a saddle point of $\mathcal{L}$ exists, one can show that iterating (\ref{4}) - (\ref{6}) converge to the solution of (\ref{2}) \cite{FixPoCon}.
The algorithm is very similar to the dual ascent and the method of multipliers, where it consists of an $\vartheta$-minimization step (\ref{4}), a $v$-minimization step (\ref{5}) and a dual variable update (\ref{6}). 
Note that the dual variable update uses a step size equal to $\rho$.

An important feature of the ADMM iterations (\ref{4})–(\ref{6}) is its modular structure. 
In particular, (\ref{4}) can be regarded as an inversion step as it involves the forward imaging model $f(x)$, whereas (\ref{5}) can be regarded as a denoising step as it involves the prior $g(v)$.

\paragraph{Inversion Step} For the forward imaging model, (\ref{4}) can be written, expanded and solved as follows,
\beqn
x^{(k+1)} := \underset{{x \in \mathbb{R}^{n}}}{\mathrm{argmin}}  \,\, ||\phi x - y||^{2}\, + \frac{\rho}{2} ||x- \tilde{x}^{(k)}||_{2}^{2}.
\eeqn
\beqn
x^{(k+1)} = \underset{x}{\mathrm{argmin}}\,\,\frac{1}{2}\bigg|\bigg| \quad
\begin{bmatrix} 
\phi  \\
\rho I 
\end{bmatrix}x \,\, -
\quad
\begin{bmatrix} 
y \\
\sqrt{\rho}\tilde{x}  
\end{bmatrix}\bigg|\bigg|^{2}.
\eeqn
\beqn
\label{eqinvimage}
x^{(k+1)} = (\phi^{T}\phi + \rho I)^{-1} (\phi^{T}y + \rho \tilde{x}).
\eeqn

\paragraph{Denoising Step} Now, we turn our attention to the \textit{denoising} step, if we can define $\omega = \sqrt{\frac{1}{\rho}} $ and represent (\ref{5}) as follows
\beqn
\label{7}
 v^{(k+1)} := \underset{v\in \mathbb{R}^{n}}{\mathrm{argmin}} \,\, \lambda g(v) + \frac{1}{2\omega^{2}} ||v- \tilde{v}^{(k)}||_{2}^{2},\\
 \label{8}
 v^{(k+1)} := \underset{v\in \mathbb{R}^{n}}{\mathrm{argmin}} \,\, \lambda g(v) + \frac{1}{2\omega^{2}} ||v- (x^{(k+1)} - \vartheta^{(k)})||_{2}^{2}.
\eeqn
Here, (\ref{8}) is a conventional \textit{image denoising} problem. 
If we treat $ x^{(k+1)} - \vartheta^{(k)}$ and $v$ as ``noisy'' image and ``clean'' image, respectively, (\ref{7}) minimizes the residue between ``noisy'' and ``clean'' image.
This can be done by using prior $g(v)$ and quadratic loss $l(v;r) = \frac{1}{2\omega^{2}} ||v - r||_{2}^{2}$, where $r = x^{(k+1)} - \vartheta^{(k)}$ at iteration $k+1$. 
Thus, we can write (\ref{8}) as 
\beqn
\label{denoise}
v^{(k+1)} := \mathcal{D}_{\omega} \bigg(x^{(k+1)} - \vartheta^{(k)}\bigg).
\eeqn
By ``denoising'', we mean recovering $v_0$ from noisy measurements $r$ of the form
\beqn
r = v_0 + e, \,\,\,\, e \thicksim \mathcal{N}(0, \sigma^{2}I),
\eeqn
for some variance $\sigma^{2} > 0$.

Many denoising algorithms, such as BM3D and non-local means, require an estimate of the noise level, which is $\omega$ for our case.
We treat it as a tunable knob to control the amount of denoising because the residual, $v- \tilde{v}^{(k)}$ at $k$th iteration is not exactly Gaussian \cite{FixPoCon}.  
Thanks to this modular structure, \cite{pnp} proposed that in the case of ADMM, one does not need to specify $g$ before running the ADMM. 
Instead, they replace (\ref{5}) by using an \textit{off-the-shelf} image denoising algorithm. 
To enhance the performance, we can develop the convolutional Denoising Autoencoder (cDAE). 
However, this paper is the first step to consider the CS framework and its results are acceptable.
Interested readers can find detailed derivations and explanations for cDAE at the online technical report \cite{Zahin19, Techreport}.

\subsection{Fall Frame Classification}
\label{ffc}

We now have only frames, in which there is an object, rather than having the large number of frames with minimal information, i.e. no object at all. 
For fall classification, we adopt a trained classifier to detect whether the person has fallen or not. 
This trained classifier is briefly presented as follows, interested readers can find the detailed steps in \cite{Fall_Action_Detection, Zahin19}.
Recall that, we reconstruct the frames at the decoder only if they have an object.
This detection step can be proceeded when we perform the background subtraction, which is described in \cite{Techreport}.
Also, we stacks all those frames and applies the binary classifier on each frame separately.

\section{Experimental Analysis}

\begin{figure}[ht]
\begin{center}
\includegraphics [width=0.7\columnwidth]{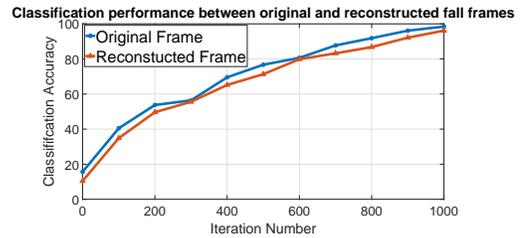}
\caption{Classification performance between original and reconstructed frames}
\label{detection_performance}
\end{center}
\end{figure}

Firstly, we select the dataset, namely the UR Fall Dataset (URFD) \cite{UCF} that is often used in the literature, which is suitable for benchmarking purposes. 
This dataset contains 30 videos of falls and 40 videos of Activities of Daily Living (ADL) (which are labeled as no falls). 
In the following, we present the necessary results, more results can be found in the technical report \cite{Techreport}.

Before comming the reconstruction, we present the denoising step. 
Owing to the larger data size, video frames need to be compressed for transmission in the wireless medium. 
Furthermore, while deploying video sensor nodes in a real environment, the captured frames can be continuously influenced by different factors, e.g. noise, channel fading, signal attenuation,  nodes' asynchronous transmission and signal time-shift phenomenon due to the spacial deployment of sensor nodes. 
To sum up, video frame degradation can be occurred from \textit{compression} and \textit{transmission} perspective. 
The results of denoised step is illustrated in Fig.~\ref{denoised_re}.

We now present the reconstruction performance.
The restored frame quality is measured, when the frame is reconstructed at the receiver/decoder side. 
The peak signal to noise ratio (PSNR) is used to evaluate the quality, where a higher PSNR value indicates better image quality. This quantity is defined as the ratio of the peak signal energy to the MSE between the recovered frame and the original frame. 
The PSNR (dB) is usually expressed as
\beqn
\label{PSNR}
PSNR = 10 \,\, log_{10}\bigg( \frac{255^2}{MSE}\bigg).
\eeqn
Fig.~\ref{reconstructed_re} demonstrates the image reconstruction performance of our proposed CS-ADMM framework. 
It took maximum 25 iterations to reach a good value according to PSNR.

Finally, we present the performance of fall-action detection.
The major contribution of our proposed framework is not only to reconstruct frames at the decoder end but also to show the classification performance with the original one. 
To this end, we show the classification performance between the original frame and the reconstructed one  in Fig.~\ref{detection_performance}. 
For this purpose, we use the already trained classifier to demonstrate the system performance.
Note that, we train the classifier with the original frames, whereas we use both the original one and the reconstructed version of those from our proposed framework for testing. 
Our simulation results show that the trained classifier obtains the comparable performance if not the best with the original one.

\section{Conclusion}

In this paper, we proposed  the novel framework for the smart health monitoring, which is the combination of CS, machine
learning based denoiser and ADMM structure. 
We furthermore focused on detecting fall down actions, where the trained binary CNN classifier is used as the fall-action classifier. 
The experimental result has shown that our framework can get the acceptable results with the low complexity. 
For further study, the experiments with video sequences are needed to study the efficacy of the proposed model. 
We also aim to use convolutional auto encoder to our proposed framework to enhance the system performance.

\end{document}